\documentclass{article}

\usepackage{arxiv}

\usepackage[utf8]{inputenc} % allow utf-8 input
\usepackage[T1]{fontenc}    % use 8-bit T1 fonts
\usepackage{hyperref}       % hyperlinks
\usepackage{url}            % simple URL typesetting
\usepackage{booktabs}       % professional-quality tables
\usepackage{amsmath}
\usepackage{amsfonts}
\usepackage{amssymb}
\usepackage{amsthm}
\usepackage{thmtools}
\usepackage{enumitem}
\usepackage{array}
\usepackage{tikz}
\usetikzlibrary{fit,shapes.geometric,backgrounds,calc}
\usepackage[round,sort&compress]{natbib}

\makeatletter
\newtheoremstyle{indented}
  {20pt}% space before
  {20pt}% space after
  {\addtolength{\@totalleftmargin}{1.em}
   \addtolength{\linewidth}{-3.5em}
   \parshape 1 2.0em \linewidth}% body font
  {}% indent
  {\bfseries}% header font
  {:}% punctuation
  {\newline}% after theorem header
  {}% header specification (empty for default)
\makeatother

\theoremstyle{indented}
\declaretheorem[style=indented]{example}

\allowdisplaybreaks[2]

\title{Assembly line balancing with task division\thanks{This paper is dedicated to the memory of Dr. Waldemar Grzechca, formerly of the Institute of Automation Control, The Silesian University of Technology, Gliwice, Poland; who passed away suddenly on June 23rd, 2015. The authors are honored to have known him and will always be grateful that he introduced to them the problem studied herein.}}

\author{
  Carlos Alexandre X. Silva\\
  \texttt{carlosalexandre@inf.ufg.br}
   \And
   Les Foulds\\
   \texttt{lesfoulds@inf.ufg.br}
   \And
   Humberto J. Longo\\
   \texttt{longo@inf.ufg.br}\\
   \And{}\\[-10pt]
  Instituto de Inform\'atica\\
  Universidade Federal de Goi\'as\\
  Alameda Palmeiras, Quadra D, Campus Samambaia.\\ CEP 74001970, Goi\^ania--GO, Brazil
}

\begin{document}
\maketitle

\begin{abstract}
In a commonly-used version of the Simple Assembly Line Balancing Problem (SALBP-1) tasks are assigned to stations along an assembly line with a fixed cycle time in order to minimize the required number of stations. It has traditionally been assumed that the total work needed for each product unit has been partitioned into economically indivisible tasks. However, in practice, it is sometimes possible to divide particular tasks in limited ways at additional time penalty cost. Despite the penalties, task division where possible, now and then leads to a reduction in the minimum number of stations. Deciding which allowable tasks to divide creates a new assembly line balancing problem, TDALBP (Task Division Assembly Line Balancing Problem). We propose a mathematical model of the TDALBP, an exact solution procedure for it and present promising computational results for the adaptation of some classical SALBP instances from the research literature. The results demonstrate that the TDALBP sometimes has the potential to significantly improve assembly line performance.
\end{abstract}

% keywords can be removed

% SECTION -----------------------
\section{Introduction}
\label{sec:intro}
Assembly line balancing is well established in industrial engineering and much research effort has been devoted over the years to analyzing one of the so-called simple assembly line balancing problems SALPB-1, which involves minimizing the required number of work stations for a given cycle time.

The purpose of the present paper is to propose a generalization of the SALPB-1 model that allows for the potential division of certain of the given tasks at the cost of additional processing time. This is an extension of the traditionally-held assumption that the total work needed for each product unit has been partitioned into economically indivisible tasks. This extension arises from practical experience in industry where it is sometimes possible to divide particular tasks in limited ways. Despite the time penalties incurred, task division may lead to a reduction in the minimum number of stations. Deciding which allowable tasks to divide creates a new assembly line balancing problem, which is here. 

The paced manufacturing assembly line for the mass production of standardized automobiles was first introduced by Ranson Eli Olds in 1901 in his Olds Motor Vehicle Company \citep{womack-jones-roos-1991}. By 1913, Henry Ford had significantly improved the assembly line concept and nowadays, the first assembly line for building cars is attributed to him. It was designed to be an efficient, highly productive way of manufacturing discrete commodities. These days, the basic assembly line is used to produce large volumes of a wide variety of standardized or customized assembled products, including not only vehicles, but also many home appliances and electronic goods. As observed by \citet{baybars-1986}, \citet{scholl-1998} and many others, the management of assembly lines is a significant short-to-medium term challenge in manufacturing planning.

Following the notation and terminology of \citet{scholl-boysen-fliedner-2009}, the line consists of a set of $m$ (work) \emph{stations} arranged in a linear fashion, with stations connected by a mechanical material handling device. The basic movement of material through an assembly line begins with the workpieces necessary to begin the creation of each unit of a single product  being fed consecutively to the first station. The work necessary to complete each unit of the product is divided into a set of \emph{tasks} $V = \{1,\dots,n\}$, each of which must be assigned to a station. Once a partially assembled product enters an activated station, an assigned (nonempty) set of tasks is performed on it and it is then fed to the next activated station.

The time required to complete each task $j \in V$ is termed its (positive, integral) \emph{processing time} and is denoted by $t_j$, $j = 1,\dots,n$. The set of tasks assigned to a station $k$ comprises its \emph{station load}, and is denoted by $S_k$. The total time required to process all the tasks assigned to each station $k = 1,\dots,m$; is termed its \emph{station time}, and is denoted by $t(S_k)$. The common time available to complete all the assigned tasks in sequence at each station is termed the \emph{cycle time}, and is denoted by $c$. Note that the station time must not exceed the cycle time at each station, i.e., $t(S_k)\leqslant c$, $k=1,\dots,m$. If $t(S_k) < c$, then station $k$ has \emph{idle time} of $(c - t(S_k))$ time units in each cycle.

There are sequences of tasks that must be followed in the assembly of any product and these are represented in a \emph{precedence graph} $G = (V, A, t)$, that has a set of arcs $A$, each representing the necessary precedence relations $(i,j)$ between different tasks $i, j \in V$. That is, task $i$ must be completed before task $j$ is commenced. $G$ is acyclic, with $V$ numbered topologically and $t:V \rightarrow \mathbb{N}$, where $t(j) = t_j$, $j = 1,\dots,n$. For each task $j \in V$, its set of direct predecessors is defined as $P_j = \{i \in V\>|\> (i, j) \in A\}$ and its set of direct successors is defined as $F_j = \{i \in V\>|\> (j, i)\in A\}$. To account for indirect precedence relationships, the set of all predecessors $P^*_j = \{i \in V\>|\>\text{a path of arcs from $i$ to $j$ exists in $G$}\}$ and the set of all successors $F^*_j = \{i \in V\>|\>\text{a path of arcs from $j$ to $i$ exists in $G$}\}$, respectively, are also defined.

The assembly line balancing problem (ALBP) involves assigning the tasks needed to produce each unit of a product among the work stations along the manufacturing line in order to optimize some given system performance measure of the line. The idea of line balancing was first introduced by \citet{bryton-1954} in his thesis as a medium term planning challenge. The first published quantitative study was reported by \citet{salveson-1955} who formulated the problem as a linear program and was the first to introduce the phrase the \textit{Assembly Line Balancing Problem}.

\citet{gutjahr-nemhauser-1964} showed that the ALBP falls into the class of $\mathcal{NP}$-hard combinatorial optimization problems. Furthermore, practical ALBP instances can be extremely large, with thousands of tasks. Thus, planners often have to examine a huge number of alternative balancing plans to deal with uncertain model mix information and technological and logistical constraints. For these reasons, heuristic methods have become popular techniques for solving the ALBP. However, with the emergence of Lagrangian relaxation and column generation, integer programming techniques are now capable of solving many practical ALBP instances. We focus on such approaches in the present paper. 

Most types of assembly line balancing problems are based on a set of limiting assumptions. \citet{baybars-1986} specified the following assumptions for the well-known simplified version of the ALBP, the Simple Assembly Line Balancing Problem (SALBP): 
\begin{enumerate}[label=A\arabic*),ref=A\arabic*]%,topsep=2pt,itemsep=2pt]
 \item\label{type-all} all input parameters are known with certainty; 
 \item\label{type-split} no task can be divided among two or more stations; 
 \item tasks cannot be processed in arbitrary sequences due to technological precedence requirements; 
 \item each task must be processed; 
 \item all stations under consideration are identically equipped and staffed to process any task; 
 \item task processing times are independent of the station at which they are processed and of the preceding tasks; 
 \item any task can be processed at any station; 
 \item the line is serial, with no parallelism nor side feeds; 
 \item the line is designed for the mass production of one model of a single product; 
 \item\label{type-c} either the cycle time $c$, is given and fixed, or 
 \item\label{type-m} the number of workstations $m$, is given and fixed.
\end{enumerate}

The most popular variants of the SALBP considered in the literature arise from assumptions \ref{type-c} and \ref{type-m} above and are: 
\begin{itemize}
 \item The Simple Assembly Line Balancing Problem-1 (SALBP-1): Given the cycle time $c$, minimize $m$, the number of stations.
 \item The Simple Assembly Line Balancing Problem-2 (SALBP-2): Given $m$, the number of stations, minimize $c$, the cycle time.
\end{itemize}

\citet{wee-magazine-1982} showed that SALBP-1 is an $\mathcal{NP}$-hard problem. Most of the above assumptions \ref{type-all}--\ref{type-m}, have been relaxed or somehow modified by various model extensions appearing in the literature. The present paper addresses the SALBP-1 variation arising when only Assumption \ref{type-split} above is relaxed to a limited extent, i.e., some tasks may potentially be divided among more than one station in particular ways at the additional cost of time penalties.

The \emph{Branch and Bound} technique has been widely used to solve SALBP-1 instances optimally. \citet{scholl-klein-1999} presented a comparison, with computational results, between approaches that use branch and bound and various other strategies to solve the SALBP-1. The SALBP-1, without the precedence constraints, can be seen as an instance of the \emph{Bin Packing Problem}, where a set of tasks (objects) has to be allocated (packed) in stations (bins) with identical finite cycle time. The solution of such an instance of the Bin Packing problem is used in many branch and bound based algorithms as a lower bound for the optimal solution of the SALBP-1, as in FABLE \citep{johnson1988optimally}, SALOME \citep{scholl-klein-1997} and BB\&R \citep{sewell-jacobson-2012}.

Exact methods for the SALBP-1 have evolved over the years with the development of computer technology and modern algorithmic techniques. The FABLE algorithm, presented by \citet{johnson1988optimally}, uses a set of lower bounds, including a solution of the Bin Packing Problem, and a series of rules of domination to discard partial solutions dominated by solutions previously explored. \citet{nourie1991finding} presented a tree structure used in their OptPack algorithm to eliminate previously exploited solutions more efficiently than the domination rules used in FABLE. \citet{hoffmann1992eureka} introduced a hybrid system, called Eureka, which uses Branch and Bound to find an optimal solution within a previously defined time limit. If this solution is not found, a heuristic is used to find an approximate solution. Hoffmann also observed that some instances could be resolved more quickly if processed in the opposite direction of the precedence graph. He proposed using half the time limit computing in one direction and the other half in the opposite direction.

\citet{scholl-klein-1997} presented the SALOME algorithm, which combined the best features of the FABLE and Eureka algorithms. In 1999 \citet{scholl-klein-1999} improved SALOME by adding a tree-based domination rule developed by \citet{nourie1991finding} for the OptPack algorithm. According to \citet{sewell-jacobson-2012}, SALOME was the best algorithm to solve SALBP-1 optimally. However, in the same article they proposed a new hybrid algorithm called BB\&R (Branch, Bound and Remember) to solve SALBP-1 optimally, with reported computational results superior to those obtained with the SALOME algorithm. The BB\&R algorithm combines a number of previously successful SALBP-1strategies including branch and bound, dynamic programming and domination rules, as reported by several authors (\cite{hoffmann1992eureka, nourie1991finding,scholl-becker-2006,scholl-klein-1997}, and others).

The remainder of the paper is organized as follows. Section \ref{sec:tdalbp} introduces a generalization of the SALBP-1 problem that allows for the limited division of certain tasks with time penalties. An optimization model for the new problem is formulated in Section \ref{sec:model-tdalbp}. In Section \ref{sec:algor} the BB\&R algorithm is detailed, as well as the necessary changes in its structure to allow the search of optimal solutions for instances of the new problem. Computational results obtained by applying the adapted algorithm to instances derived from classical SALBP-1 instances of the literature are reported in Section \ref{sec:results}. Finally, a conclusion with some proposals for future research are given in Section \ref{sec:conclusions}.

% SECTION -----------------------
\section{The Task Division Assembly Line Balancing Problem ({\small TDALBP})}
\label{sec:tdalbp}

It is traditionally assumed in assembly line balancing that the total work needed for each unit of any product has been partitioned into economically indivisible tasks, so that further division would create unnecessary subtasks and an unproductive increase in total work content. Thus, as stated in Assumption \ref{type-split} in Section \ref{sec:intro}, each required task is indivisible, in the sense that it must be performed at a single station. In practice in industry, however, this is not always so. Changes in technology, product design, manufacturing processes and job design may create opportunities to make beneficial changes to the way in which the total work content is partitioned into tasks. This may mean that some of the so-called indivisible tasks become potentially divisible in the sense that they can be profitably divided further into subtasks that are assigned to different stations. This may be especially the case when there is a time-oriented objective, such as in the SALBP-1, and there is significant variance among task processing times with some of them equal to, or close to, the cycle time. When some of the tasks with relatively longer processing times involve simple operations such as: boring, drilling, or tightening bolts or screws, part of the task might be usefully delayed to a later station. For example, suppose a task consists of tightening many bolts and this task has one of the longer processing times among all tasks. Suppose that only two bolts need to be tightened to hold two workpieces safely in place and the rest of the bolts can be tightened at a later station, without access problems caused by the completion of intermediate tasks. In such a case, the minimal number of required stations may possibly be reduced when a task like this can be divided in such way, even when the division induces additional time penalty costs. This will be illustrated in a numerical example given later.

Including the possibility of such task division leads to a new assembly line balancing problem that we denote by TDALBP (Task Division Assembly Line Balancing Problem). This concept has been discussed in qualitative ways by a number of authors (c.f., for example, \citet{chase-jacobs-2013}). But, as far as we are aware, apart from a brief introduction to the problem by \citet{grzecha-foulds-2015b}, the issue has not been discussed in the open literature in a quantitative manner. It is the purpose of the present paper to attempt to fill this gap in the literature by discussing a rigorous version of the problem, formulating a binary integer programming model of it, proposing solution methods and reporting computational experience.

The TDALBP is a generalization of the SALBP-1 with Assumption \ref{type-split} above relaxed to a limited, specific extent in the sense that a given subset of tasks can potentially be divided in some particular, known ways. Instead of confining each such task to be processed at exactly one station, it can potentially be divided by assigning each its consequent subtasks to different stations, with the sum of the processing times of the subtasks equal to the processing time of the original task. Any such division may causes additional time penalties to be incurred at any station to which a subtask is assigned.

We formulate the TDALBP in a manner similar to how the ASALBP (alternative subgraphs assembly line balancing problem) was formulated by \citet{scholl-boysen-fliedner-2009}. The ASALBP extends the SALBP by relaxing the additional tacit assumption that all tasks are processed in a predetermined mode and no processing alternatives exist. Instead, the authors assume that alternatives exist for parts of the production process. Each such part is called a variable subprocess which results in alternative (disjunctive) subgraphs in the precedence graph of which exactly one has to be chosen for each part.

In our formulation of the TDALBP we also assume that processing alternatives exist, but only for a given subset of potentially divisible tasks. The node representing each potentially divisible task in the precedence graph is expanded to a set of additional nodes (each with the same precedence relations as the original node) where each new node represents a possible subtask created by dividing the original task. 

It is theoretically possible to consider the TDALBP as a special case of the ASALBP in which each potentially divisible task is considered to be a variable subprocess and each feasible combination of its subtasks is represented by an alternative subgraph. In this way, any TDALBP instance can be solved by any ASALBP method, such as those proposed by \citet{capacho-pastor-dolgui-guschinskaya-2009} and \citet{scholl-boysen-fliedner-2009}. The difficulty is that in the TDALBP, each potentially divisible task may possibly be divided into many subtasks. Furthermore, feasible subtask combinations do not correspond to only partitions of the processing time, as illustrated in the following example. These factors may lead in practice to an unmanageable number of ASALBP subgraphs.

\begin{example}
\label{ex:1}
Suppose a certain TDALBP instance has a particular potentially divisible task with processing time 6. The task can be divided into subtasks with the following processing times: \{6\}, \{5, 1\}, \{4, 2\}, \{4, 1, 1\}, \{3, 3\}, \{3, 2, 1\}, \{3, 1, 1, 1\}, \{2, 1, 1, 1, 1\} and \{1, 1, 1, 1, 1, 1\}. Note that choosing the first ``subtask'' option reflects the decision to not divide the task. Exactly one of these combinations must be chosen. If the task is actually divided into two or more subtasks, its subtasks are all assigned to different stations. In such a case, the processing of each subtask incurs a station-independent time penalty (to be introduced later.)
\end{example}

Thus, if no restrictions are placed on the division process, it is clear that the number of division options grows exponentially with the number of given subtasks. This leads to a huge number of ASALBP alternative subgraphs for all but trivial TDALBP instances. For this reason, we prefer to construct a dedicated TDALBP model that contains only the set of given task division combination options themselves, whose cardinality grows in $\mathcal{O}(1)$. Identifying optimal combinations is left to solution algorithms for the model. 

We now propose a version of the TDALBP that allows for just a single combination of subtasks for each potentially divisible task to be chosen. This leads to the TDALBP optimization model given in the next section. The node representing each potentially divisible task in $V$, is expanded to include a new set of nodes in $G$ that represent all subtasks that could be created if the task is divided. Each new node is allocated the same precedence relations as the original node. The following notation is introduced to make the presentation more rigorous. Let:
\begin{description}[align=right,labelwidth=!,leftmargin=4.3cm,labelsep=.15cm]
 \item [$D$ :] the set of potentially divisible tasks $(D \subseteq V, D \neq \emptyset)$;
 \item [$I$ :] the set of indivisible tasks $(I \subseteq V, n \in I, V = D \cup I, D \cap I = \emptyset)$;
 \item [$r_j$ :] the number of processing times for which task $j$ can be processed, $\forall\> j \in V$ ($r_j$ is set as unity, $\forall\> j \in I$);
 \item [$t_j^q$ :] the (positive, integral) $q^\text{th}$ processing time available for task $j$, $\forall\> j \in V,\> q = 1,\dots,r_j$;
 \item [$T_j = \langle t_j^q\>|\>q = 1,\dots,r_j\rangle$ :] the (nonincreasing) sequence of given processing times for which task $j \in V$ can be processed, where $t_j^1 = t_j$, the original processing time of the task $j$ and thus $t_j^p \geqslant t_j^q$ whenever $p < q$, for $1\leqslant p, q \leqslant r_j$ ($T_j$ contains only the single entry $t_j^1$, $\forall\> j \in I$);
 \item [$f_j^q :$] the (strictly positive) time penalty incurred if task $j$ is processed for $t_j^q$ time units, $\forall\> j\in V$, $q = 1,\dots,r_j$ ($f_j^1$ is set as zero for each original task $j$, $\forall\> j \in V$).
\end{description}

With only a slight abuse of terminology, from now on, we use the terms tasks and nodes interchangeably. It is assumed that the terminal node $n$, is the only one in $G$ without successors. If this does not occur naturally in $G$, a dummy last node with zero processing time is introduced. It is also assumed that task $n$ is indivisible. To enable certain possibly useful subtask combinations of each potentially divisible task $j \in D$ to be considered, repetitions of subtask processing times $t_j^q$, for certain values of $q$, $2 \leqslant q \leqslant r_j$, may appear in $T_j$, allowing $j$ to be divided into subtasks with equal processing times. However, it is assumed that such subtasks must still be processed at separate stations. For instance, if $t_j = 6$, it may be useful to divide j into two subtasks to be processed at different stations, each with given processing time of 3 units. There is no time penalty if task $j$ is not divided and which is why $f_j^1$ is defined as zero. If potentially divisible task $j$ is divided, and this results in a subtask with a given processing time of $t_j^q$ being activated at any station, the final processing time of the subtask is set to $(t_j^q + f_ j^q)$, $\forall \> j\in D, q = 1,\dots,r_j$. Note that all $f_j^q$ values are station-independent. Furthermore, it is assumed that there are no precedence relations between subtasks of the same potentially divisible task.

The TDALBP consists of activating a subset of subtasks for each potentially divisible task $j \in D$, by assigning them to stations (where the task $j$ itself is considered a subtask) with the following characteristics:
\begin{itemize}
 \item the activated subtasks all satisfy the precedence relations involving task $j$,
 \item the sum of the given processing times of the activated subtasks equals the given processing time $t_j$, of task $j$,
 \item the total processing time of each activated divided subtask is the sum of its given processing and penalty times, and
 \item each activated divided subtask of task $j$ must be processed at a distinct station, $\forall\> j \in D$.
\end{itemize}

The subtasks of each potentially divisible task that are not activated and their incident arcs in $G$ are ignored. The objective of the TDALBP is to identify the optimal set of subtasks of all potentially divisible tasks so that all activated tasks (the indivisible tasks and the subtasks of potentially divisible tasks) can be assigned to the minimal possible number of stations while taking account their final processing times, the given precedence constraints and the cycle time.
 
Some observations on conditions for well-formulated TDALBP instances are in order. If the given processing time of any task exceeds the cycle time, it must be divisible as a necessary condition for the existence of a feasible solution. As described in the next section, there is a standard approach to establishing the set of assignable stations to which any task can be assigned. If the number of subtasks in a particular subtask combination of a potentially divisible task $j$, exceeds the number of potentially assignable stations, the combination can be discarded.

To illustrate this point, consider Example \ref{ex:1}, given earlier. One of the subtask combinations is $\{1, 1, 1, 1, 1, 1\}$. If there are at most five stations to which the parent potentially divisible task can be assigned, this combination (which requires six distinct stations) can be discarded.

Also, any subtask of a potentially divisible task $j$ with a given processing time that renders it impossible to be part of a subtask combination whose given processing times sum to $t_j$ can also be discarded. Furthermore, as all the time penalties $t_j^q$, $j\in D$, $q=2,\dots,r_j$, are assumed to be strictly positive, there is no advantage in dividing a potentially divisible task $j$, into a combination of subtasks that contains a subtask with given processing time $(t_j^q - 1)$, for some $2 \leqslant q \leqslant r_j$. Thus, it is assumed henceforth that the given processing times for all subtasks (apart from task $j$ itself) $t_j^q$, are such that $1 \leqslant t_j^q \leqslant (t_j - 2)$; $j \in D, q = 2,\dots,r_j$. A numerical example of the TDALBP is provided as Example \ref{ex:2} below. In the next Section we formulate a mathematical model of the TDALBP.

\begin{example}
\label{ex:2}
A TDALBP instance is given in Figure \ref{fig:2} and Table \ref{tab:1}. The potentially divisible tasks: $D=\{2, 3, 9, 13, 14, 18\}$, are depicted by round nodes and the set of indivisible tasks: $V\backslash D=\{1, 4, 5, 6, 7, 8, 10, 11, 12, 15, 16, 17, 19,\allowbreak 20, 21, 22, 23\}$, are depicted by square nodes in Figure \ref{fig:1}. The original given task processing times are shown above the nodes. All feasible subtasks and their given processing and penalty times for the potentially divisible tasks are shown in Table \ref{tab:1}.

\begin{figure}[htbp]
 \caption{The precedence graph of the numerical example.}
 \label{fig:1}
 \centering
\begin{tikzpicture}[
  ->,>=stealth,%shorten >=1pt,
  auto,node distance=2.0cm,thick,
  every node/.style={rectangle,draw,
                                 fill=red!10,
                                 inner sep=0pt,
                                 align=center,
                                 minimum size=8mm,
                                 font=\sffamily\bfseries\normalsize},
  scale=.70, transform shape
]
 \node (n1) [label={[label distance=-5pt]90:$5$}] {$1$};
 \node (n2) [circle,fill=green!10,below of=n1,label={[label distance=-5pt]90:$6$}] {$2$};
 \node (n3) [circle,fill=green!10,right of=n1,label={[label distance=-5pt]90:$7$}] {$3$};
 \node (n4) [right of=n2,label={[label distance=-5pt]90:$2$}] {$4$};
 \node (n5) [right of=n3,yshift=-1.1cm,label={[label distance=-5pt]90:$5$}] {$5$};
 \node (n7) [right of=n5,label={[label distance=-5pt]90:$1$}] {$7$};
 \node (n6) [above of=n7,yshift=.8cm,label={[label distance=-5pt]90:$3$}] {$6$};
 \node (n8) [below of=n7,label={[label distance=-5pt]90:8}] {$8$};
 \node (n9) [circle,fill=green!10,right of=n6,yshift=1cm,label={[label distance=-5pt]90:$8$}] {$9$};
 \node (n10) [below of=n9,label={[label distance=-5pt]90:$6$}] {$10$};
 \node (n11) [right of=n7,label={[label distance=-5pt]90:$4$}] {$11$};
 \node (n12) [right of=n8,label={[label distance=-5pt]90:$2$}] {$12$};
 \node (n13) [circle,fill=green!10,right of=n9,label={[label distance=-5pt]90:$5$}] {$13$};
 \node (n14) [circle,fill=green!10,right of=n10,label={[label distance=-5pt]90:$8$}] {$14$};
 \node (n15) [right of=n11,label={[label distance=-5pt]90:$2$}] {$15$};
 \node (n16) [right of=n12,label={[label distance=-5pt]90:$3$}] {$16$};
 \node (n17) [right of=n13,yshift=-1cm,label={[label distance=-5pt]90:$1$}] {$17$};
 \node (n18) [circle,fill=green!10,right of=n15,yshift=-1cm,label={[label distance=-5pt]90:$8$}] {$18$};
 \node (n19) [right of=n13,xshift=2cm,yshift=-.3cm,label={[label distance=-5pt]90:$3$}] {$19$};
 \node (n20) [right of=n14,xshift=2cm,yshift=+.3cm,label={[label distance=-5pt]90:$3$}] {$20$};
 \node (n21) [right of=n15,xshift=2cm,yshift=-.3cm,label={[label distance=-5pt]90:$2$}] {$21$};
 \node (n22) [right of=n16,xshift=2cm,yshift=+.3cm,label={[label distance=-5pt]90:$2$}] {$22$};
 \node (n23) [right of=n20,yshift=-1.3cm,label={[label distance=-5pt]90:$6$}] {$23$};
  \path (n1) edge (n3)
           (n2) edge (n4)
           (n3) edge [bend left=10] (n5)
           (n4) edge [bend right=10] (n5)
           (n5) edge [bend left=10] (n6)
           (n5) edge (n7)
           (n5) edge [bend right=10] (n8)
           (n6) edge [bend left=10] (n9)
           (n6) edge [bend right=10] (n10)
           (n7) edge (n11)
           (n8) edge (n12)
           (n9) edge (n13)
           (n10) edge (n14)
           (n11) edge (n15)
           (n12) edge (n16)
           (n13) edge [bend left=10] (n17)
           (n14) edge [bend right=10] (n17)
           (n15) edge [bend left=10] (n18)
           (n16) edge [bend right=10] (n18)
           (n17) edge [bend left=10] (n19)
           (n17) edge [bend right=10] (n20)
           (n18) edge [bend left=10] (n21)
           (n18) edge [bend right=10] (n22)
           (n19) edge [bend left=10] (n23)
           (n20) edge [bend left=10] (n23)
           (n21) edge [bend right=10] (n23)
           (n22) edge [bend right=10] (n23);
\end{tikzpicture}
\end{figure}
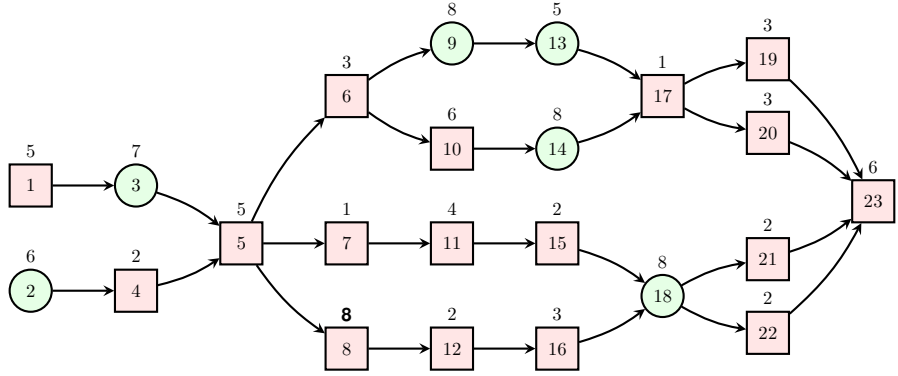
 \begin{table}[htbp]
 \centering
\caption{Input data for the TDALBP instance shown in Figure \ref{fig:1}.}
\label{tab:1}
% \resizebox{\textwidth}{!}{
  \begin{tabular}{cccccccccccc}
   \hline
$j$ & $t_j^1$ & $t_j^2$ & $t_j^3$ & $t_j^1 + f_j^1$ & $t_j^2 + f_j^2$ & $t_j^3 + f_j^3$ & $f_j^1$ & $f_j^2$ & $f_j^3$ & $E_j$ & $F_j$\\
   \hline
  2 & 6 & 3 & 3 & 6 & 4 & 4 & 0 & 1 & 1 & 2 & 3\\
  3 & 7 & 4 & 3 & 7 & 5 & 4 & 0 & 1 & 1 & 1 & 3\\
  9 & 8 & 6 & 2 & 8 & 7 & 3 & 0 & 1 & 1 & 4 & 10\\
13 & 5 & 3 & 2 & 5 & 4 & 3 & 0 & 1 & 1 & 5 & 11\\
14 & 8 & 5 & 3 & 8 & 6 & 4 & 0 & 1 & 1 & 4 & 10\\
18 & 8 & 6 & 2 & 8 & 7 & 3 & 0 & 1 & 1 & 5 & 11\\
   \hline
  \end{tabular}
% }
 \end{table}

As can be seen in the table, each potentially divisible task $j$, has $r_j = 3$ processing options: $T_j = \langle t_j^1, t_j^2, t_j^3\rangle$. Note that two subtasks of one potentially divisible task have the same given processing times ($t_2^2 = t_2^3 = 3$). In this example, all time penalties are equal to unity. Figure \ref{fig:2} displays the expanded version of $G$ that includes all of the possible subtasks and their final processing times. To make the graph transformation process clear, the task indices have not been renumbered topologically.
\begin{figure}[htbp]
 \caption{The expanded version of the precedence graph of the numerical example.}
 \label{fig:2}
 \centering
\begin{tikzpicture}[
  ->,>=stealth,%shorten >=1pt,
  auto,node distance=2.0cm,thick,
  every node/.style={rectangle,draw,
                                 fill=red!10,
                                 inner sep=0pt,
                                 align=center,
                                 minimum size=8mm,
                                 font=\sffamily\bfseries\normalsize},
  scale=.6, transform shape
]
 \node (n1) [label={[label distance=-5pt]90:$5$},label={[label distance=2pt,blue!80!black]-90:$S_2$}] {$1$};
 \node (n3-2) [circle,fill=green!10,right of=n1,label={[label distance=-5pt]90:$5$}] {$3^2$};
 \node (n3-1) [circle,fill=green!10,above of=n3-2,yshift=-.5cm,label={[label distance=-5pt]90:$7$}] {$3^1$};
 \node (n3-3) [circle,fill=green!10,below of=n3-2,yshift=.5cm,label={[label distance=-5pt]90:$4$}] {$3^3$};
 \node (n2-1) [circle,fill=green!10,below left of=n3-3,xshift=-.6cm,label={[label distance=-5pt]90:$6$},label={[label distance=1pt,blue!80!black]30:$S_1$}] {$2^1$};
 \node (n2-2) [circle,fill=green!10,below of=n2-1,yshift=.5cm,label={[label distance=-5pt]90:$4$}] {$2^2$};
 \node (n2-3) [circle,fill=green!10,below of=n2-2,yshift=.5cm,label={[label distance=-5pt]90:$4$}] {$2^3$};
 \node (n4) [right of=n2-2,label={[label distance=-5pt]90:$2$}] {$4$};
 \node (n5) [right of=n3-3,yshift=-1.5cm,label={[label distance=-5pt]85:$5$},label={[label distance=3pt,blue!80!black]180:$S_3$}] {$5$};
 \node (n7) [right of=n5,label={[label distance=-5pt]90:$1$}] {$7$};
 \node (n6) [above of=n7,yshift=5.5cm,label={[label distance=-5pt]90:$3$},label={[label distance=2pt,blue!80!black]180:$S_4$}] {$6$};
 \node (n8) [below of=n7,label={[label distance=-5pt]90:$8$},label={[label distance=-4pt,blue!80!black]235:$S_5$}] {$8$};
 \node (n9-2) [circle,fill=green!10,right of=n6,label={[label distance=-5pt]90:$7$}] {$9^2$};
 \node (n9-1) [circle,fill=green!10,above of=n9-2,yshift=-.5cm,label={[label distance=-5pt]90:$8$}] {$9^1$};
 \node (n9-3) [circle,fill=green!10,below of=n9-2,yshift=.5cm,label={[label distance=-5pt]90:$3$},label={[label distance=-3pt,blue!80!black]235:$S_6$}] {$9^3$};
 \node (n10) [below of=n9-3,yshift=-1cm,label={[label distance=-5pt]90:$6$},label={[label distance=-4pt,blue!80!black]235:$S_7$}] {$10$};
 \node (n14-2) [circle,fill=green!10,right of=n10,label={[label distance=-5pt]90:$6$}] {$14^2$};
 \node (n14-1) [circle,fill=green!10,above of=n14-2,yshift=-.5cm,label={[label distance=-5pt]90:$8$},label={[label distance=-4pt,blue!80!black]135:$S_8$}] {$14^1$};
 \node (n14-3) [circle,fill=green!10,below of=n14-2,yshift=.5cm,label={[label distance=-5pt]90:$4$}] {$14^3$};
 \node (n11) [right of=n7,label={[label distance=-5pt]90:$4$}] {$11$};
 \node (n12) [right of=n8,label={[label distance=-5pt]90:$2$}] {$12$};
 \node (n13-1) [circle,fill=green!10,right of=n9-1,label={[label distance=-5pt]90:$5$}] {$13^1$};
 \node (n13-2) [circle,fill=green!10,right of=n9-2,label={[label distance=-5pt]90:$4$}] {$13^2$};
 \node (n13-3) [circle,fill=green!10,right of=n9-3,label={[label distance=-5pt]90:$3$}] {$13^3$};
 \node (n15) [right of=n11,label={[label distance=-5pt]90:$2$}] {$15$};
 \node (n16) [right of=n12,label={[label distance=-5pt]90:$3$}] {$16$};
 \node (n17) [right of=n13-3,yshift=-1cm,label={[xshift=2pt,label distance=-5pt]87:$1$}] {$17$};
 \node (n18-2) [circle,fill=green!10,right of=n15,label={[label distance=-5pt]90:$7$}] {$18^2$};
 \node (n18-1) [circle,fill=green!10,above of=n18-2,label={[label distance=-5pt]90:$8$},label={[label distance=-4pt,blue!80!black]60:$S_{10}$}] {$18^1$};
 \node (n18-3) [circle,fill=green!10,below of=n18-2,label={[label distance=-5pt]90:$3$}] {$18^3$};
 \node (n19) [right of=n17,yshift=1cm,label={[label distance=-3pt]90:$3$},label={[label distance=-5pt,blue!80!black]50:$S_9$}] {$19$};
 \node (n20) [right of=n17,yshift=-1cm,label={[label distance=-5pt]90:$3$}] {$20$};
 \node (n21) [right of=n18-2,yshift=1cm,label={[label distance=-5pt]90:$2$}] {$21$};
 \node (n22) [right of=n18-2,yshift=-1cm,label={[label distance=-5pt]90:$2$}] {$22$};
 \node (n23) [right of=n20,yshift=-1.5cm,label={[xshift=2pt,label distance=-5pt]87:$6$},label={[label distance=9pt,blue!80!black]180:$S_{11}$}] {$23$};

  \path (n1) edge [bend left=10] (n3-1)
           (n1) edge (n3-2)
           (n1) edge [bend right=10] (n3-3)
           (n2-1) edge [bend left=10] (n4)
           (n2-2) edge (n4)
           (n2-3) edge [bend right=10] (n4)
           (n3-1) edge [bend left=10] (n5)
           (n3-2) edge [bend left=10] (n5)
           (n3-3) edge [bend left=10] (n5)
           (n4) edge [bend right=10] (n5)
           (n5) edge [bend left=10] (n6)
           (n5) edge (n7)
           (n5) edge [bend right=10] (n8)
           (n6) edge [bend left=10] (n9-1)
           (n6) edge (n9-2)
           (n6) edge [bend right=10] (n9-3)
           (n6) edge [bend right=10] (n10)
           (n7) edge (n11)
           (n8) edge (n12)
           (n9-1) edge (n13-1)
           (n9-1) edge [bend right=10] (n13-2)
           (n9-1) edge [bend right=10] (n13-3)
           (n9-2) edge [bend left=10] (n13-1)
           (n9-2) edge (n13-2)
           (n9-2) edge [bend right=10] (n13-3)
           (n9-3) edge [bend left=10] (n13-1)
           (n9-3) edge [bend left=10] (n13-2)
           (n9-3) edge (n13-3)
           (n10) edge [bend left=10] (n14-1)
           (n10) edge (n14-2)
           (n10) edge [bend right=10](n14-3)
           (n11) edge (n15)
           (n12) edge (n16)
           (n13-1) edge [bend left=10] (n17)
           (n13-2) edge [bend left=10] (n17)
           (n13-3) edge [bend left=10] (n17)
           (n14-1) edge [bend right=10] (n17)
           (n14-2) edge [bend right=10] (n17)
           (n14-3) edge [bend right=10] (n17)
           (n15) edge [bend left=10] (n18-1)
           (n15) edge (n18-2)
           (n15) edge [bend right=10] (n18-3)
           (n16) edge [bend right=10] (n18-1)
           (n16) edge [bend right=10] (n18-2)
           (n16) edge (n18-3)
           (n17) edge [bend left=10] (n19)
           (n17) edge [bend right=10] (n20)
           (n18-1) edge [bend left=10] (n21)
           (n18-1) edge [bend right=10] (n22)
           (n18-2) edge [bend left=10] (n21)
           (n18-2) edge [bend right=10] (n22)
           (n18-3) edge [bend left=10] (n21)
           (n18-3) edge [bend right=10] (n22)
           (n19) edge [bend left=10] (n23)
           (n20) edge [bend left=10] (n23)
           (n21) edge [bend right=10] (n23)
           (n22) edge [bend right=10] (n23);

\begin{scope}[on background layer]
\fill[blue!20,rounded corners=4mm]
  ([xshift=-2mm,yshift=2mm]n2-1.north west) --
  ([xshift=+2mm,yshift=2mm]n2-1.north east) --
  ([xshift=+2mm,yshift=2mm]n4.north east) --
  ([xshift=+2mm,yshift=-2mm]n4.south east) --
  ([xshift=-2mm,yshift=-2mm]n4.south west) --
  ([xshift=-2mm,yshift=2mm]n4.west) --
  ([xshift=2mm,yshift=-2mm]n2-1.south east) --
  ([xshift=-2mm,yshift=-2mm]n2-1.south west) --
  cycle %S1
  ([xshift=-2mm,yshift=2mm]n1.north west) --
  ([xshift=+2mm,yshift=2mm]n3-2.north east) --
  ([xshift=+2mm,yshift=-2mm]n3-2.south east) --
  ([xshift=-2mm,yshift=-2mm]n1.south west) --
  cycle %S2
  ([xshift=-2mm,yshift=2mm]n3-3.north west) --
  ([xshift=+2mm,yshift=2mm]n3-3.north east) --
  ([xshift=+2mm,yshift=2mm]n5.north east) --
  ([xshift=+2mm,yshift=-2mm]n5.south east) --
  ([xshift=-2mm,yshift=-2mm]n5.south west) --
  ([xshift=-2mm,yshift=2mm]n5.west) --
  ([xshift=2mm,yshift=-2mm]n3-3.south east) --
  ([xshift=-2mm,yshift=-2mm]n3-3.south west) --
  cycle %S3
  ([xshift=-2mm,yshift=2mm]n6.north west) --
  ([xshift=+2mm,yshift=2mm]n9-2.north east) --
  ([xshift=+2mm,yshift=-2mm]n9-2.south east) --
  ([xshift=-2mm,yshift=-2mm]n6.south west) --
  cycle %S4
  ([xshift=-2mm,yshift=2mm]n8.north west) --
  ([xshift=+2mm,yshift=2mm]n12.north east) --
  ([xshift=+2mm,yshift=-2mm]n12.south east) --
  ([xshift=-2mm,yshift=-2mm]n8.south west) --
  cycle %S5
  ([xshift=-2mm,yshift=2mm]n7.north west) --
  ([xshift=-1mm,yshift=-7mm]n9-3.south) --
  ([xshift=-2mm,yshift=-1mm]n9-3.south west) --
  ([xshift=-3mm,yshift=2mm]n9-3.north west) --
  ([xshift=2mm,yshift=1mm]n9-3.north) --
  ([xshift=-1mm,yshift=-2mm]n13-1.south west) --
  ([xshift=-3mm,yshift=2mm]n13-1.north west) --
  ([xshift=3mm,yshift=2mm]n13-1.north east) --
  ([xshift=2mm,yshift=-2mm]n13-1.south east) --
  ([xshift=-2mm,yshift=-2mm]n13-1.south) --
  ([xshift=4mm,yshift=1mm]n9-3.north) --
  ([xshift=3mm,yshift=-2mm]n9-3.south east) --
  ([xshift=-1mm,yshift=6mm]n7.north) --
  ([xshift=2mm,yshift=2mm]n7.north east) --
  ([xshift=2mm,yshift=-2mm]n7.south east) --
  ([xshift=-2mm,yshift=-2mm]n7.south west) --
  cycle %S6
  ([xshift=-2mm,yshift=2mm]n10.north west) --
  ([xshift=2mm,yshift=2mm]n10.north east) --
  ([xshift=2mm,yshift=-2mm]n10.south east) --
  ([yshift=-2mm]n10.south west) --
  ([xshift=-5mm]n11.west) --
  ([xshift=-5mm,yshift=-3mm]n11.south west) --
  ([yshift=-5mm]n11.south east) --
  ([xshift=-2mm]n16.north west) --
  ([xshift=2mm,yshift=2mm]n16.north west) --
  ([xshift=2mm,yshift=2mm]n16.north east) --
  ([xshift=2mm,yshift=-2mm]n16.south east) --
  ([xshift=-2mm,yshift=-2mm]n16.south west) --
  ([xshift=-2mm]n16.west) --
  ([yshift=-8mm]n11.south east) --
  ([xshift=-5mm,yshift=-5mm]n11.south west) --
  ([xshift=-7mm]n11.west) --
  ([xshift=-2mm,yshift=-2mm]n10.south west) --
  cycle %S7
  ([xshift=-2mm,yshift=2mm]n14-1.north west) --
  ([xshift=+2mm,yshift=3mm]n17.north east) --
  ([xshift=+2mm,yshift=-2mm]n17.south east) --
  ([xshift=-2mm,yshift=-3mm]n14-1.south west) --
  cycle %S8
  ([xshift=-2mm,yshift=2mm]n11.north west) --
  ([xshift=2mm,yshift=2mm]n11.north) --
  ([xshift=-6mm]n14-3.west) --
  ([xshift=-1mm,yshift=3mm]n14-3.north west) --
  ([xshift=-4mm,yshift=-2mm]n20.south west) --
  ([xshift=-2mm,yshift=2mm]n19.north west) --
  ([xshift=2mm,yshift=2mm]n19.north east) --
  ([xshift=2mm,yshift=-2mm]n20.south east) --
  ([xshift=-2mm,yshift=-2mm]n20.south west) --
  ([xshift=-1mm,yshift=0mm]n14-3.north west) --
  ([xshift=-2mm]n14-3.west) --
  ([xshift=1mm,yshift=4mm]n11.north east) --
  ([xshift=2mm,yshift=-2mm]n11.south east) --
  ([xshift=-2mm,yshift=-2mm]n11.south west) --
  cycle %S9
  ([xshift=-4mm,yshift=1mm]n15.north west) --
  ([yshift=1mm]n15.north) --
  ([xshift=-2mm,yshift=-2mm]n18-1.west) --
  ([xshift=-2mm,yshift=3mm]n18-1.north west) --
  ([xshift=+3mm,yshift=2mm]n18-1.north east) --
  ([xshift=+2mm,yshift=-2mm]n18-1.south east) --
  ([xshift=2mm,yshift=2mm]n15.east) --
  ([xshift=2mm,yshift=-2mm]n15.south east) --
  ([xshift=-2mm,yshift=-3mm]n15.south west) --
  cycle %S10
  ([xshift=-2mm,yshift=3mm]n21.north west) --
  ([xshift=2mm,yshift=1mm]n21.north) --
  ([xshift=-3mm,yshift=-3mm]n23.west) --
  ([xshift=-2mm,yshift=2mm]n23.north west) --
  ([xshift=+2mm,yshift=2mm]n23.north east) --
  ([xshift=+3mm,yshift=-1mm]n23.south east) --
  ([xshift=1mm,yshift=-5mm]n23.south) --
  ([xshift=+2mm,yshift=1mm]n22.east) --
  ([xshift=+2mm,yshift=-2mm]n22.south east) --
  ([xshift=-2mm,yshift=-2mm]n22.south west) --
  cycle %S11
 ;
\end{scope}
\end{tikzpicture}
\end{figure}
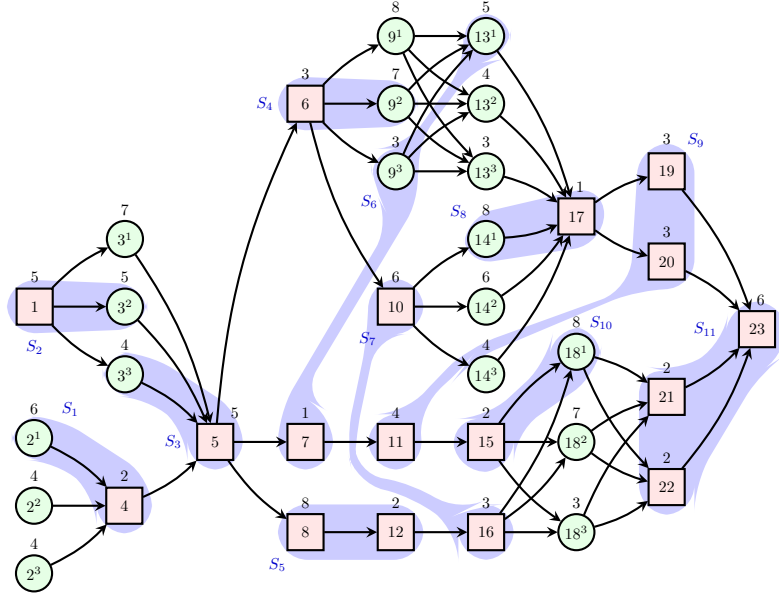

This TDALBP can be initially treated as an SALBP-1 instance, with no division permitted and hence with no time penalties. One optimal solution to this instance with a given cycle time of $c = 10$, has station loads: $S_1 = \{1\}$, $S_2 = \{2, 4\}$, $S_3 = \{3\}$, $S_4 = \{5, 6, 7\}$, $S_5 = \{8, 12\}$, $S_6 = \{9\}$, $S_7 = \{10, 11\}$, $S_8 = \{13, 15, 16\}$, $S_9 = \{18, 22\}$, $S_{10} = \{14, 21\}$, $S_{11} = \{17, 19, 20\}$ and $S_{12} = \{23\}$, with a minimal number of stations, $m^* =12$. The obvious question to ask now is, considering the example as a TDALBP, is it possible to improve on this solution by permitting the potential task divisions given in Table \ref{tab:1}, along with their associated time penalties? An optimal solution to this problem, still with $c = 10$, has station loads: $S_1 = \{1, 3^2\}$, $S_2 = \{2, 3^3\}$, $S_3 = \{4, 5, 7\}$, $S_4 = \{6, 9^2\}$, $S_5 = \{8, 12\}$, $S_6 = \{9^3, 11, 16\}$, $S_7 = \{10, 14^3\}$, $S_8 = \{13^2, 14^2\}$, $S_9 = \{15, 18\}$, $S_{10} = \{13^3, 17, 19, 20\}$, $S_{11} = \{21, 22, 23\}$, with a minimal number of stations, $m^* = 11$. (Note that it is feasible to process subtask $14^3$ before subtask $14^2$ as there are no precedence relations between subtasks of the same potentially divisible task.) Furthermore, of the six potentially divisible tasks in $D = \{2, 3, 9, 13, 14, 18\}$, four are actually divided (tasks 3, 9, 13 and 14) and two are not divided (tasks 2 and 18). Despite an additional total time penalty of $f_3^2 + f_3^3 + f_9^2 + f_9^3 + f_{13}^2 + f_{13}^3 + f_{14}^2 + f_{14}^3 = 8$ time units, the minimal number of stations has been reduced by one from the earlier solution to 11.
 
Another optimal solution to this problem, still with the minimal number of stations equal to 11, but with an additional total time penalty of only $f_3^2 + f_3^3 + f_9^2 + f_9^3 = 4$ time units, has station loads: $S_1 = \{2, 4\}$, $S_2 = \{1, 3^2\}$, $S_3 = \{3^3, 5\}$, $S_4 = \{6, 9^2\}$, $S_5 = \{8, 12\}$, $S_6 = \{7,9^3, 13\}$, $S_7 = \{10, 16\}$, $S_8 = \{14, 17\}$, $S_9 = \{11,19, 20\}$, $S_{10} = \{15, 18\}$, $S_{11} = \{21, 22, 23\}$. The sets $S_i$, $i=1,\dots, 11$, relating to this solution, are depicted in Figure \ref{fig:2}.
 
A further optimal solution with the minimal additional total time penalty of 2 ($f_3^2 + f_3^3$) time units, has station loads: $S_1 = \{2, 4\}$, $S_2 = \{1, 3^2\}$, $S_3 = \{3^3, 5,7\}$, $S_4 = \{6,10\}$, $S_5 = \{8\}$, $S_6 = \{9, 12\}$, $S_7 = \{11,15,16\}$, $S_8 = \{18, 22\}$, $S_9 = \{14, 21\}$, $S_{10} = \{13, 17,19\}$, $S_{11} = \{20, 23\}$.
\end{example}

% SECTION -----------------------
\section{An optimization model for the TDALBP}
\label{sec:model-tdalbp}

We now present a relatively compact model of the TDALBP. It is based in part on the previous SALBP-1 models of \citet{patterson-albracht-1975} and of \citet{scholl-1998}; and the ASALBP models of \citet{capacho-pastor-dolgui-guschinskaya-2009} and of \citet{scholl-boysen-fliedner-2009}.

As mentioned earlier, task $n$ is assumed indivisible and is to be assumed unique in not having any successor tasks. Each potentially divisible node $j$ is used to generate a node set $D_j = \{j\}\cup\{j^q\>|\>q = 2,\dots,r_j\}$, of the given possible subtasks of task $j$, (including j itself) corresponding to the entries in $T_j$, i.e.: $t_j^q$, $\forall\> j \in D, q = 1,\dots,r_j$. Each node $j \in D$ is replaced by $D_j$ in $G$. Each new node $j^q$, in the expanded version of $G$ is allocated the same precedence relations as those of node $j$, $\forall\> j \in D, q = 2,\dots,r_j$. This is achieved in the following manner. If arc $(h^p, j^q)\in A$ for some $p$, where $1 \leqslant p \leqslant r_h$; and for some $j^q$, where $j \in D$ and $1 \leqslant q \leqslant r_j$; then arcs $(h^p, j^q)$, $\forall\> p = 1,\dots,r_h$; and $q = 1,\dots,r_j$; are added to $A$. If arc $(j^q, i^s)\in A$ for some $j^q$, where $j \in D$ and $1 \leqslant q \leqslant r_j$; and for some $s$, where $1 \leqslant s \leqslant r_i$; then arcs $(j^q, i^s)$, $\forall\> q = 1,\dots,r_j$; and all $s = 1,\dots,r_i$; are added to $A$. Each new node $j^q$ is allocated a revised processing time of $(t_j^q + f_j^q)$. The processing times of all the original nodes in $V$ are unchanged. Thus $V$ is now partitioned into $V = I \cup (\cup_{j\in D} D_j)$ and $n$ is increased to reflect the new total number of nodes. 

Each indivisible task in $I$ must be assigned to exactly one station. For each potentially divisible node $j\in D$, consider the subtasks in $D_j$, that include task $j$ itself. Either: ($i$) task $j$ alone is activated, undivided and is assigned to a single station where it is processed in its given processing time $t_j$, without time penalty, or ($ii$) one or more subtasks of $D_j$ (excluding task $j$) are activated and assigned to separate stations where each is processed in its given time $t_j^q$, plus a further time penalty, $f_j^q$. The processing times of the activated subtasks must sum to the given processing time $t_j$, of task $j$. All the nodes in the transformed graph are re-indexed topologically. For each potentially divisible node $j$, a record must be kept of its newly-created subtasks nodes in $D_j$, $\forall\> j \in D$. The transformational process is illustrated in Example 2 in Figures \ref{fig:1} and \ref{fig:2}. Strictly speaking, all nodes should be re-indexed topologically in order to use the proposed model in the next section. However, so that the transformation process is clear, topological re-indexing was not carried out, with the labels of the new nodes left in the style $j^q$.
 
In order to make the proposed model as compact as possible, the following notation and terminology are introduced for the expanded graph. For each task $j$ let $E_j$ be the earliest station to which $j$ can be feasibly assigned, given the precedence relations, the given processing times and the cycle time, $\forall\> j \in V$. The usual formula for calculating $E_j$ for the SALBP-1 was first introduced by \citet{patterson-albracht-1975}. Recall that for the TDALBP, the sequence $T_j$, of given processing times of the possible subtasks of each potentially divisible task $j$ is strictly decreasing and thus $t_j^{r_j}$ is minimal among all elements of $T_j$, $\forall\> j \in D$. Using this fact, the following slight modification of the Patterson and Albracht formula can be used to calculate $E_j$ for the ASALBP:
\begin{equation}\label{eq:Ej}
E_j = 
\begin{cases}
1, & \text{if }(t_j^{r_j} + \sum_{i \in P^*_j} t_i) = 0,\text{ and}\\
\lceil(t_j^{r_j} + \sum_{i \in P^*_j} t_i) / c\rceil, & \text{otherwise}, \forall\> j \in V.
\end{cases}
\end{equation}

Letting $L_j$ be the latest station to which each task $j$ can be feasibly assigned, given the precedence relations, the given processing times and the cycle time, $\forall\> j \in V$. Then $L_j$ can be calculated analogously as:
\begin{equation}
L_j =
\begin{cases}\label{eq:Fj}
 m', & \text{if }(t_j^{r_j} + \sum_{i \in F^*_j} t_i) = 0, \text{ and}\\
 m' + 1 - \lceil(t_j^{r_j} + \sum_{i \in F^*_j} t_i) \rceil, &\text{otherwise}, \forall\>j \in V,
\end{cases}
\end{equation}
where $m'$ ($ \leqslant n$) is the maximum number of stations necessary to identify a feasible solution.

Note that $m'$ is often computed by a heuristic algorithm, but if this is unavailable, $m'$ can be set to be $n$, implying that initially, each task is assigned to its own individual station. Once the parameters defined in \eqref{eq:Ej} and \eqref{eq:Fj} are established, the \textit{station interval} $SI_j = [E_j, F_j]$, can be computed which defines the interval of stations in which task $j$ can be assigned. If task $j$ is indivisible (potentially divisible or a subtask) it must be assigned to exactly (at most one station) in $SI_j$, $\forall\>j \in V$. To further enhance the compactness of the model to be proposed, we introduce the \textit{station interval} $B_k = \{j \in V\>|\> k \in SI_j\}$, for each station $k = 1,\dots,m'$. Use of the $SI_j$ and $B_k$ parameters enable a significant reduction (compared with the traditional models of other SALBP variations) in the number of constraints to be achieved in the proposed model. The model, which is based in part on the ASALBP model of \citet{scholl-boysen-fliedner-2009}, is now introduced. In the description of the model the following indices, parameters and variables are used:
\begin{itemize}
 \item Indices:
  \begin{description}[align=right,labelwidth=!,leftmargin=2.3cm,labelsep=.3cm]
   \item [$j\sim\!\!$] for tasks,
   \item [$k\sim\!\!$] for stations and
   \item [$q\sim\!\!$] for subtasks and their given processing times.
  \end{description}
 \item Parameters:
  \begin{description}[align=right,labelwidth=!,leftmargin=2.3cm,labelsep=.3cm]
   \item [$n$ :\!\!] the number of tasks ($\forall\>j = 1,\dots,n$),
   \item [$m'$ :\!\!] the maximum required number of stations,% ($\forall\>k = 1,\dots,m'$),
   \item [$V$ :\!\!] the set of all tasks ($|V| = n$),
   \item [$I$ :\!\!] the set of indivisible tasks,
   \item [$D$ :\!\!] the set of potentially divisible tasks,
   \item [$D_j$ :\!\!]  the set of subtasks (including $j$ itself) into which the potentially divisible task $j$ can be divided ($\forall\>j \in D$),
   \item [$t_j$ :\!\!] the given processing time (without a time penalty) for task $j$ ($\forall\>j = 1,\dots,n$),
   \item [$f_j$ :\!\!]  the time penalty for processing task $j$ ($\forall\> j = 1,\dots,n$),
   \item [$P_j\> (F_j)$ :\!\!] the set of immediate predecessors (successors) of task $j$ ($\forall\>j = 1,\dots,n$),
   \item [$E_j$ :\!\!] the earliest station to which task $j$ can be feasibly assigned ($\forall\>j = 1,\dots,n$),
   \item [$L_j$ :\!\!] the latest station to which task $j$ can be feasibly assigned ($\forall\>j = 1,\dots,n$),
   \item [$SI_j = {[}E_j, L_j{]}$ :\!\!] the interval of stations to which task $j$ can be assigned ($\forall\>j = 1,\dots,n$),
   \item [$B_k$ :\!\!] the set of tasks that can be potentially assigned to station $k$ ($\forall\> k = 1,\dots,m'$).
  \end{description}
 \item Decision variables:
  \begin{description}[align=right,labelwidth=!,leftmargin=2.3cm,labelsep=.3cm]
   \item [$x_{jk}^q=\!\!$]
        $\begin{cases}
           1,&\text{if subtask $q$ of task $j$ is assigned to station $k$};\\
	   0,&\text{otherwise; }\> j = 1,\dots,n;\> q = 1,\dots,r_j;\> \forall\> k \in SI_j.
	  \end{cases}$
  \end{description}
\end{itemize}

The station index to which task $j$ is assigned, if it is assigned at all, must be a member of $SI_j$ and can be computed as $\sum_{q=1}^{r_j}\sum_{k \in SI_j} k{\cdot}x_{jk}^q$, $\forall\>j = 1,\dots,n$. If this expression turns out to be zero for a given set of decision variables $x_{jk}^q$, then task $j$ is unassigned. 

If a task $j \in I$ (or a subtask $q$ of $j \in D$) is activated at some station $\kappa$, then $x_{j\kappa}^1 = 1$ (or $ x_{j\kappa}^q = 1$). The $\kappa$ index is given by:
\begin{align}
 \kappa = \sum_{k \in SI_j} k{\cdot} x_{jk}^q, \> j = 1, \dots, n \label{eq:index}.
\end{align}
If $\kappa = 0$, then task $j$ (or subtask $q$ of $j$) has not been activated. Note that only tasks that are members of $B_k$ can be assigned to station $k$, for $k = 1, \dots, m'$.

As will be seen in \eqref{eq:tdalbp-1} below, because it is assumed that the terminal node in $G$ has no successors, if $j$ is set to $n$ in the previous expression, the required number of stations can be expressed as a simple weighted sum. Furthermore, only tasks that are members of $B_k$ can be assigned to station $k$, $\forall\>k= 1,\dots,m'$. The proposed model, \eqref{eq:tdalbp-1}--\eqref{eq:tdalbp-10}, which involves only the binary decision variables $x_{jk}^q$, is given next. If task $j$ is indivisible, that is $j\in I$ and $r_j=1$, then $x_{jk}^1$ and $t_j^1$ are denoted simply by $x_{jk}$ and $t_j$, respectively.

\begin{alignat}{2}
 \omit\rlap{\underline{\bf Model TDALBP}:} \nonumber\\[10pt]
 \omit\rlap{Minimize $\displaystyle\sum_{k\in SI_n}k{\cdot}x_{nk}$,}\label{eq:tdalbp-1}\\
 \omit\rlap{subject to}\nonumber\\
 \sum_{k\in SI_j} x_{ik}                                         & = 1,    &\> & i \in I; \label{eq:tdalbp-2}\\
 \sum_{k\in SI_j}\sum_{q\in D_j}t_j^q{\cdot}x_{jk}^q & = t_j^1,   & & j\in D;\label{eq:tdalbp-3}\\
 \sum_{q\in D_j}x_{jk}^q                                         & \leqslant 1, & & j\in D,\nonumber\\[-15pt]
 & & & k \in SI_j;\label{eq:tdalbp-4}\\
 \sum_{i\in I\cap B_k} t_i{\cdot}x_{ik} + \sum_{j\in D}\sum_{q\in D_j\cap B_k}(t_j^q+f_j^q){\cdot}x_{jk}^q & \leqslant c, & & k=1,\dots,m';\label{eq:tdalbp-5}\\
 \sum_{k\in SI_i}k{\cdot}x_{ik} - \sum_{k\in SI_j}k{\cdot}x_{jk} & \leqslant 0, & & i\in I,\nonumber\\[-15pt]
 & & & j \in I\cap F_i,\nonumber\\
 & & & L_i\geqslant E_j;\label{eq:tdalbp-6}\\
 \sum_{k\in SI_i}k{\cdot}x_{ik} - \sum_{k\in SI_j}k{\cdot}x_{jk}^q + m'{\cdot}\sum_{k\in SI_j}x_{jk}^q & \leqslant m', & & q \in \mathop{\bigcup}_{j\in D}D_j\cap F_i,\nonumber\\
 & & & i\in I,\nonumber\\
 & & & L_i\geqslant E_j;\label{eq:tdalbp-7}\\
 \sum_{k\in SI_i}k{\cdot}x_{ik}^q - \sum_{k\in SI_j}k{\cdot}x_{jk} + m'{\cdot}\sum_{k\in SI_i}x_{ik}^q & \leqslant m', & & q\in \mathop{\bigcup}_{i\in D}D_i,\nonumber\\
  & & & j \in I\cap F_i,\nonumber\\
  & & & L_i\geqslant E_j;\label{eq:tdalbp-8}\\
 \sum_{k\in SI_{i}}k{\cdot}x_{ik}^{q'} - \sum_{k\in SI_{j}}k{\cdot}x_{jk}^{q''} + m'{\cdot}(\sum_{k\in SI_{i}}x_{ik}^{q'} + \sum_{k\in SI_{j}}x_{jk}^{q''}) & \leqslant 2{\cdot}m', & & q'\in \mathop{\bigcup}_{i\in D}D_{i},\nonumber\\
  & & & q'' \in \mathop{\bigcup}_{j\in D}D_{j}\cap F_{i},\nonumber\\
  & & & L_{i}\geqslant E_{j};\label{eq:tdalbp-9}\\
 x_{jk}^q & \in \{0, 1\}, & & q=1,\dots,r_j,\nonumber\\
  & & & j=1,\dots,n,\nonumber\\
  & & & k\in SI_j.\label{eq:tdalbp-10}
\end{alignat}

The objective \eqref{eq:tdalbp-1} involves minimizing the station index of the terminal task $n$, which is equivalent to minimizing the number of stations required. Constraints \eqref{eq:tdalbp-2} ensure that each indivisible task is assigned to exactly one station. Constraints \eqref{eq:tdalbp-3} guarantee that the sum of the given processing times of the activated subtasks of each potentially divisible task equals the processing time of the task. Satisfying constraint \eqref{eq:tdalbp-4} ensures that the activated subtasks of any potentially divisible task are each assigned to a separate station. The cycle time constraints are given in \eqref{eq:tdalbp-5}, where, for each station,  the first expression represents the processing times of the indivisible tasks (without time penalties)  and the second represents those of the potentially divisible tasks (with time penalties). \eqref{eq:tdalbp-6}--\eqref{eq:tdalbp-9} deal with possible task precedence constraints that may arise from the existence of arc $(i, j) \in A$, depending on the nature of, and relationships between the tasks $i$ and $j$. (Recall that the set of subtasks of any potentially divisible task include the task itself.) When $i$ and $j$ are both indivisible, the precedence constraints are given in \eqref{eq:tdalbp-6}. When $i$ is indivisible but $j$ is a subtask of a potentially divisible task, the precedence constraint arising from $(i, j)$ must hold only if $j$ is activated. That is, when $j$ is assigned to some station, i.e., $\sum_{k \in SI_j} x_{jk}^q = 1$. The corresponding constraints are given in \eqref{eq:tdalbp-7}. When $i$ is a subtask of a potentially divisible task but $j$ is indivisible, the precedence constraint arising from $(i, j)$ must hold only if $i$ is activated. That is, when $i$ is assigned to some station, i.e., $\sum_{k \in SI_i} x_{ik} = 1$. The corresponding constraints are given in \eqref{eq:tdalbp-8}. When both $i$ and $j$ are subtasks of two distinct potentially divisible tasks, the precedence constraint arising from $(i, j)$ must hold only if both $i$ and $j$ are activated. That is, when $i$ and $j$ are both assigned to stations, i.e., $\sum_{k \in SI_i} x_{ik} = \sum_{k \in SI_j} x_{jk}^q = 1$. The corresponding constraints are given in \eqref{eq:tdalbp-9}. Constraints \eqref{eq:tdalbp-10} are the usual binary conditions on the $x_{jk}^q$'s.

Compared to the model SBF of \citet{scholl-boysen-fliedner-2009}, the numbers of binary variables are reduced significantly, because SBF includes (up to) $m'{\cdot}(n + |D|)$ assignment variables, while the new model requires (up to) $m'{\cdot}n$ assignment variables. However, SBF and the new model have about the same number of constraints. Using the $\mathcal{O}$-notation, the order of magnitudes are $\mathcal{O}((n + |D|)^2 + m')$ and $\mathcal{O}(|I|{\cdot}|D| + m')$ constraints for SBF and the new model, respectively. Utilizing the relations $m' \leqslant n$, $|I| \leqslant n$ and $|D| \leqslant n$, both models require $\mathcal{O}(n^2)$ variables and constraints.

% SECTION -----------------------
\section{A BB\&R based algorithm for the TDALBP}
\label{sec:algor}
First, we outline the hybrid branch, bound and remember (BB\&R) algorithm for the SALBP-1, as proposed by \citet{sewell-jacobson-2012}. The basic principle of the algorithm is the enumeration of partial solutions for the SALBP-1. A partial solution is a part of a feasible solution, and may have tasks not yet assigned to the stations. However, the tasks already assigned respect the cycle time and precedence constraint. In the enumeration process, partial solutions are extended by the assignment of free tasks to stations that still have idle time. A partial solution is discarded if it has already been generated before, or if its number of stations is greater than the number of stations of any viable solution, or if it is dominated by another partial solution. This process is repeated for each partial solution until a viable solution is found, that is, until all tasks are assigned  to stations (see the following notation).

A partial solution is denoted as $\mathcal{E} = (A, U, S_1, \dots, S_m)$, where: 
\begin{description}[align=right, labelwidth=!, leftmargin=2.4cm,labelsep=.2cm]
	\setlength\itemsep{1em}
  \item [$m$ :] number of stations used by $\mathcal{E}$;
  \item [$S_k$ :] set of tasks assigned to station $k$ in $\mathcal{E}$, $k=1,\dots,m$;
  \item [$A = \bigcup\limits_{k=1}^m S_k$ :] set of tasks assigned to the $m$ stations; and
  \item [$U$ :] set of tasks still not assigned to any station.
\end{description}

The SALBP-1 without the precedence constraints reduces to a Bin Packing Problem, where a set of tasks (items) must be allocated (stored) to stations with identical cycle time (bin size). Thus, a lower bound for the SALBP-1 is obtained by the solution of a such instance of the Bin Packing Problem. \citet{sewell-jacobson-2012} proposed the Bin Packing Algorithm (BPA), which is based on a previous algorithm proposed by \citet{korf-2003}, to obtain a tighter lower bound called BINLB. The BB\&R algorithm also uses the three following standard lower bounds \citep{scholl-klein-1997,scholl-becker-2006}:
\begin{align}
	LB_1 & = \left\lceil\sum\nolimits_{j \in U}\frac{t_j}{c}\right\rceil;\label{eq:LB1BBR}\\ 
    LB_2 & = \left\lceil\sum\nolimits_{j \in U} w_j^1\right\rceil;\\
    LB_3 & = \left\lceil\sum\nolimits_{j \in U} w_j^2\right\rceil;
    \intertext{where}
    w_j^1 & =
    	\begin{cases}
        	1, & \text{if } t_j > \frac{c}{2}, \\
        	\frac{1}{2}, & \text{if } t_j = \frac{c}{2};\\
      	\end{cases}
    \intertext{and}
   	w_j^2 & =
        \begin{cases}
        	1, & \text{if } t_j > \frac{2{\cdot}c}{3}, \\
        	\frac{2}{3}, & \text{if } t_j = \frac{2{\cdot}c}{3}, \\
            \frac{1}{2}, & \text{if }\ \frac{c}{3} < t_j < 			\frac{2{\cdot}c}{3}, \\
        	\frac{1}{3}, & \text{if } t_j = \frac{c}{3}.
        \end{cases}
\end{align}

To decrease the number of subproblems that are generated after the branching, \citet{sewell-jacobson-2012} used a heuristic to obtain an upper bound. This heuristic, denoted as MHH (modified Hoffmann heuristic), is a modified version of the well known heuristic proposed by \citet{hoffmann-1963} for the SALBP-1.

The BB\&R algorithm uses the function in \eqref{eq:expand} to decide which is the best partial solution to be expanded:
\begin{equation}
\mathcal{B}(\mathcal{E}) = \mathcal{I}/m - \lambda|U|;\label{eq:expand}
\end{equation}
where $\mathcal{E} = (A, U, S_1, ..., S_m)$ is a partial solution, $\mathcal{I}$ is the total idle time in all the $m$ stations and $\lambda$ is a balancing factor (\citet{sewell-jacobson-2012} used the value $\lambda = 0.002$ which was determined empirically through preliminary computational tests with a subset of instances). Sewell and Jacobson argue that the term $\mathcal{I}/m$ induces full or nearly full loads. When the other values are equal, $-\lambda|U|$ induces loads that contain fewer tasks, leaving smaller tasks to the end, making it easier to assign these tasks to the stations.

The main search strategy used in the BB\&R algorithm is a CBFS search (Cyclic Best-First Search), initially called Distributed Best-First Search by \citet{Kao-2008}, that combines Best-First Search and DFS (Depth First Search). The CBFS starts expanding the best partial solution at level 1 of the search tree, then expands the best partial solution of the level 2, and so on, until the deepest level of the search tree is reached, then it cycles to the level 1 and continues. A minimum priority queue is created  for each level of the search tree and is used to select the partial solution to be expanded at the level, using  the $\mathcal{B}$ function defined in \eqref{eq:expand} as the selection criteria. 

The BB\&R algorithm with the CBFS search (BB\&R-CBFS algorithm) is divided into three steps. In step I the MHH is run to find an initial upper bound. In step II BB\&R uses the CBFS search strategy. The objectives of the second stage are to find a solution and prove its optimality, if possible. Some limits are imposed in order to avoid spending large amounts of time and memory trying to find a load for a station. The maximum load generated by a partial solution is limited to $10{,}000$ and the maximum size of each priority queue is set to $300{,}000$. So, the step II proves optimality if none of these limits is exceeded. In step III, which is only executed when step II cannot prove optimality, a BFS (Breadth First Search) strategy is used. The goal of this step is to prove the optimality of the best solution found in the previous steps.

We now discuss how parts of the BB\&R-CBFS algorithm were adapted in order to construct a TDALBP algorithm. The new algorithm, denoted here BB\&R-TD, has the same structure as the BB\&R-CBFS algorithm. In fact, the behavior of both algorithms are very similar because most of the techniques used in them are station-oriented and not-task oriented, for example, the branching and the searching strategies. The domination rules used are either station-oriented or can be applied directly to BB\&R-TD without any changes being required. \citet{sewell-jacobson-2012} used the three following domination rules, which can also be used in the BB\&R-TD algorithm. These rule are not memory-based, i.e., a partial solution does not have to be directly compared with a previously processed one. The rules are: ($i$) Maximum Load Rule -- if a partial solution contains a station that does not have a full load then it can be pruned. ($ii$) Extended Jackson's rule -- a task $i$ potentially dominates a task $j$, if there is no precedence relationship between $i$ and $j$, $t_i \geqslant t_j$ and $F^*_i \supseteq F^*_j$. If $t_i = t_j$ and $F^*_i = F^*_j$, then $i$ dominates $j$ when $i < j$. If the task $i$ dominates $j$ and $j$ is assigned to a station $s$, in which the task $i$ can also be assigned (task $i$ has not been assigned to any other station and its assignment does not violate the cycle time), then $i$ must be assigned before $j$. ($iii$) BB\&R rule -- Let $\mathcal{E} = (A, U, S_1, ..., S_m)$ be a partial solution with the $m$ stations with full load, where none of the tasks assigned to the station $ S_m $ have successors. Given an unassigned task that has successors, let $\mathcal{E}'$ be a solution created from $\mathcal{E}$, and $S_{m'}$, with $m' > m$, be the first station that has a task with a successor in $ \mathcal{E}'$. Then the loads at the stations $ S_m $ and $Sm'$ can be exchanged without violating any rule of precedence. Thus, $\mathcal{E}$ can be pruned.

The only required changes to the BB\&R algorithm relate to the calculation of bounds on the value of the optimal solution.  When task division is allowed, some lower bounds and strategies previously described do not apply directly to the TDALBP. To be valid for the TDALBL, these bounds and strategies must be modified to take into account not only individual subtask processing times but also time penalties.

The lower bound $LB_1$ is calculated as specified in \eqref{eq:LB1BBR} and the bounds $LB_2$ and $LB_3$ are now computed as follows:
\begin{align}
    	LB_2 &= \left\lceil\sum_{j \in U \cap I} w_j^1 + \sum_{j \in U \cap D}\sum_{q = 2}^{r_j} w_{j^q}^1\right\rceil\\
    	LB_3 &= \left\lceil\sum_{j \in U \cap I} w_j^2 + \sum_{j \in U \cap D}\sum_{q = 2}^{r_j} w_{j^q}^2\right\rceil
\end{align}  

In the TDALBP the task set $V$ also contains the subtasks of each potentially divisible task and is partitioned into $V = I \cup (\cup_{j\in D} D_j)$. Thus, a partial solution to the  BB\&R-TD is expanded to $\mathcal{E} = (A, U, Z, S_1, \dots, S_m)$, where $Z$ contains the number of subtasks, of each task $j\in D$ that was assigned to a station. In the enumeration process, when considering adding a divisible task $j$ to the current partial solution, a check is made to see if any of its subtasks has already been assigned to a station, that is if $Z[j] > 0$. Likewise, when a subtask $i\in D_j$ is processed, wether or not the original task $j$ is not assigned to a station is checked, i.e. if $j \not\in A$. Thus, the penalties need not be considered in the enumeration process. Also, after processing all tasks and arriving at a solution, it can be ensured that one of the two following alternatives occurs, either $(i)$ a potentially divisible task is totally assigned to a station and no one of its subtasks are assigned to stations, or $(ii)$ only the subtasks of a potentially divisible task are assigned to stations.

It was also necessary to define a new test to check the optimality of a solution to the TDALBP. Let $S_a = \min\{S_k\mid 1\leqslant k\leqslant m\}$ 
%$S_a = \{S_a \leqslant S_k\mid k \in W \}$
be the station that has the lowest processing load over all stations, $p(S_k)$ be the sum of tasks penalties in the station $k$  
%be the sum of the station $k$ task penalties
and $T_{oc} =\sum_{1\leqslant k \leqslant m} [c - (t(S_k) - p(S_k))]$ 
%$T_{oc} =\sum_{k \in W} c - (t(S_k) - p(S_k))$ 
be the sum of the idle time of all stations, neglecting of time penalties. If $T_{oc} < t(S_a)$ then the load of $S_a$ is greater than the sum of the idle time of all other stations. Thus, dividing the tasks in $S_a$, even without time penalty costs, does not help as it would be impossible to assign them to other stations. So, if this test is positive, the process can be terminated.

% SECTION -----------------------
\section{Computational experience}
\label{sec:results}

In order to evaluate the potential of the TDALBP model,  we performed computational experiments based on test instances that were constructed systematically. The instances used in the tests were generated from a set of 269 SALBP-1 instances, based on 25 precedence graphs with 8 to 297 nodes, with different cycle times. The detailed description of this data set is available at \cite{albp-2016}. To generate different types of TDALBP instances, we adapted each SALBP-1 instance, maintaining the same cycle time, tasks and precedence, but introduced the parameters $I$, $D$, $D_j$, $r_j$ and $f_j$ defined earlier and specific values for them. The nodes in the precedence graph associated with each generated instance were renumbered in topological order. Two methods were defined and tested to select the set $D$ and to create the subtasks for each task in it:
\begin{description}
 \item[\bf Method-M:] To select the set $D$ of potentially divisible tasks, we used a simple heuristic that evaluates the median of the task execution times ($X(t_j)$). The tasks that have a processing time greater than the median were selected to construct $D$ and the remaining tasks comprised $I$ (the set of indivisible tasks). $D$ was then partitioned into two subsets, $D_1$ and $D_2$ such that a task $j\in D$ was allocated to $D_1$ if $t_j > \delta{\cdot}X(t_j)$ or was allocated to $D_2$ if $t_j \leqslant \delta{\cdot}X(t_j)$. The values $\delta=1{,}2$ and $\delta = 1{,}5$ were tested. The tasks in $D_1$ were partitioned into two subtasks and those in $D_2$ were partitioned into three subtasks.
  \item[\bf Method-R:] $30\%$ of the totality of tasks in $V$ were selected, in a random way, to construct the set $D$ of potentially divisible tasks. Each of the tasks $j\in D$  was partitioned into two subtasks with a probability of $60\%$ or into three subtasks, with a probability of $40\%$.
\end{description}
In both methods the sum of the execution times of the subtasks of a particular task is always equal to that of the original task. An additional cost of one time unit was assigned to each subtask, as a time penalty.

\subsection{Results with instances analyzed by Method-M}
\label{ssec:results-M}
Table \ref{tab:2} displays the results of tests with the model TDALBP model, implemented via the CPLEX package, where the instances were generated by Method-M. A limit of $1{,}800$ seconds of processing time was imposed on each instance. The table shows only the instances where a reduction in the required number of stations was achieved. The table also contains two measures commonly used to evaluate the quality of a feasible solution: Line Efficiency (LE), which is the percentage of utilization of the line; and Line Time (LT), which is the period of time required for each unit of a product to be completed on the assembly line.

The following parameters and measurements are presented in Table \ref{tab:2}: $c$ -- cycle time; $n$ -- number of tasks; $m^*$ -- required number of stations; $\text{LE}$ -- line efficiency; $\text{LT}$ -- line time, for the ALBP solution. They are also shown for the TDALBP solution constructed with parameters $\delta=1.2$ and $\delta=1.5$, along with $F$ the total time penalty imposed. The last column, ``Obs.'', indicates whether or not the processing time limit of $1{,}800$ seconds was reached in solving the TDALBP instances (the value 1 indicates that the time limit was reached for both $\delta=1{.}5$ and $\delta=1{.}2$, the value 2 point out that the time limit was reached only for $\delta=1{.}5$ and, similarly, the values 3 refers only to $\delta=1{.}2$).
%(1 -- indicates that the time limit was reached for both $\delta=1{.}5$ and $\delta=1{.}2$, 2 -- only for $\delta=1{.}5$ and 3 -- only for $\delta=1{.}2$).

\begin{table}[htbp]
 \centering\small\setlength{\tabcolsep}{4pt}
\caption{Computational results obtained with model TDALBP and Method-M instances.}
\label{tab:2}
\resizebox{\textwidth}{!}{
  \begin{tabular}{>{\scriptsize}rlccccrccrcrccrcrc}
   \hline
   & \multicolumn{6}{c}{SALBP-1} & & \multicolumn{4}{c}{TD ($\delta = 1{.}5$)} & & \multicolumn{4}{c}{TD ($\delta = 1{.}2$)}\\
   \cline{2-7}\cline{9-12}\cline{14-17}
	&	Graph	&	$c$	&	$n$	&	$m^*$	&	LE	&	LT	&	&	$m^*$	&	$F$	&	$\text{LE}$	&	$\text{LT}$	&	&	$m^*$	&	$F$	&	$\text{LE}$	&	$\text{LT}$	&	Obs.	\\
\hline
1	&	GUNTHER	&	41	&	35	&	14	&	84.15	& 	539	&	&	13	&	9	&	90.62 	&	532	&	&	13	&	9	&	90.62	&	532	&		\\
2	&	SAWYER30	&	30	&	30	&	12	&	90.00	& 	353	&	&	11	&	3	&	98.18	&	328	&	&	11	&	3	&	98.18	&	328	&		\\
3	&	WEE-MAG	&	56	&	75	&	30	&	89.23	& 	1676	&	&	28	&	14	&	96.49	&	1568	&	&	28	&	26	&	97.26	&	1568	&	1	\\
4	&	WEE-MAG	&	54	&	75	&	31	&	89.55	&	1669	&	&	30	&	22	&	92.53	&	1615	&	&	30	&	25	&	92.53	&	1620	&	1	\\
5	&	WEE-MAG	&	45	&	75	&	38	&	87.66	&	1663	&	&	35	&	18	&	95.17	&	1573	&	&	35	&	21	&	95.17	&	1575	&	1	\\
6	&	WEE-MAG	&	43	&	75	&	50	&	69.72	&	2134	&	&	39	&	34	&	89.39	&	1669	&	&	39	&	28	&	89.39	&	1670	&		\\
7	&	WEE-MAG	&	42	&	75	&	55	&	64.89	&	2310	&	&	41	&	36	&	89.14	&	1716	&	&	40	&	32	&	91.13	&	1679	&	2	\\
8	&	WEE-MAG	&	41	&	75	&	59	&	61.97	&	2419	&	&	42	&	34	&	87.05	&	1719	&	&	42	&	38	&	87.05	&	1717	&		\\
9	&	WEE-MAG	&	35	&	75	&	60	&	71.38	&	2096	&	&	45	&	42	&	97.84	&	1575	&	&	46	&	35	&	95.28	&	1610	&	1	\\
10	&	WEE-MAG	&	36	&	75	&	60	&	69.40	&	2151	&	&	44	&	36	&	94.63	&	1584	&	&	44	&	44	&	94.63	&	1584	&	1	\\
11	&	WEE-MAG	&	37	&	75	&	60	&	67.52	&	2219	&	&	43	&	38	&	96.61	&	1589	&	&	44	&	43	&	94.72	&	1626	&	1	\\
12	&	WEE-MAG	&	38	&	75	&	60	&	65.75	&	2267	&	&	43	&	38	&	91.74	&	1631	&	&	43	&	39	&	91.74	&	1634	&	1	\\
13	&	WEE-MAG	&	39	&	75	&	60	&	64.06	&	2334	&	&	42	&	38	&	91.51	&	1634	&	&	42	&	36	&	91.51	&	1638	&		\\
14	&	WEE-MAG	&	40	&	75	&	60	&	62.46	&	2381	&	&	42	&	38	&	89.23	&	1678	&	&	42	&	38	&	89.23	&	1677	&		\\
15	&	WEE-MAG	&	33	&	75	&	61	&	74.47	&	2008	&	&	57	&	40	&	81.82	&	1875	&	&	57	&	30	&	81.29	&	1876	&	3	\\
16	&	WEE-MAG	&	34	&	75	&	61	&	72.28	&	2073	&	&	53	&	34	&	85.07	&	1796	&	&	53	&	48	&	85.85	&	1800	&		\\
17	&	WARNECKE	&	58	&	58	&	29	&	92.03	&	1679	&	&	---	&	---	&	---	&	---	&	&	28	&	27	&	96.98	&	1618	&	3	\\
18	&	WARNECKE	&	54	&	58	&	31	&	92.47	&	1672	&	&	---	&	---	&	---	&	---	&	&	30	&	30	&	97.41	&	1618	&	3	\\
   \hline
  \end{tabular}
}
 \end{table}
 
Despite the simplicity Method-M being used to select the potentially divisible tasks, the results show the potential of task division in optimizing the performance of an assembly line balancing system. The value chosen for the parameter $\delta$ directly influences the final results, changing the way that tasks are subdivided. With $\delta=1{.}5$, a reduction in the number of stations was achieved in 16 of the 269 instances generated. With $\delta=1{.}2$, in addition to these 16 instances, a reduction in the number of stations was achieved in two further instances. However, there are variations in the total amount of time penalty incurred, line efficiency, line time and even in the number of stations in the TDALBP solutions, both for better for worse. This can be seen on lines 7, 9 and 11 of the table. Note that in these three cases the execution time limit has been reached and a better solution may exist for those instances.

\subsection{Results with instances analyzed by Method-R}
We ran both the CPLEX package and the BB\&R-TD algorithm on each of the 269 TDALBP instances generated using Method-R, with a time limit of $3{,}600$ seconds for each. It was possible to identify 37 instances having a lesser number of required stations than for the SALBP-1 solutions. On the other hand, for a further set of 31 instances the processing time limit was reached without an improvement of the SALBP-1 solution, and in the remaining 201 instances the optimal solutions turned out to have the same value as that for the SALBP-1 solutions.

Method-R was then used to generate two new versions of each of these 37 successful instances, with different sets $D$ of divisible tasks. Table \ref{tab:resultsRandom} shows only the best results obtained. The first block of columns of the table, labeled ``SALBP-1'', describe the instance precedence graph, the cycle time and the required number of stations for the SALBP-1 instance. The next two blocks provide information concerning the execution of the CPLEX package and the BB\&R-TD algorithm, respectively, for each TDALB instance. In each block, the columns describe, in sequence, the required number of stations, the total time penalty imposed, the line efficiency, the line time and the time spent by the method to find the TDALBP solution.

The column labeled ``Obs.'' lists indices of some comments about the best solutions found by the CPLEX package and the BB\&R-TD algorithm. All the results were obtained within the processing time limit of $3{,}600 seconds$. The meaning of each number is (1) both solutions are optimal and were obtained from the same version of the instance; (2) both solutions are optimal, but were obtained from different versions of the instance; (3) both solutions were obtained from the same version of the instance, but only the BB\&R-TD solution is optimal; (4) the solutions were obtained from different versions of the instance and only the BB\&R-TD solution is optimal; (5) the solutions were obtained from different versions of the instance and only the CPLEX solution is optimal; and (6) the solutions were obtained from different versions of the instance but neither of them is optimal.

The results with Method-R show that for each instance, in at least one of the three versions tested, the best solution found has a reduced number of required stations,  compared to the SALBP-1 solution. The solutions of some instances, for example those of indexes 36 and 37, have line efficiency (LE) of less than 90\%. These instances still have the potential for an even greater reduction in the required number of stations, which might be achieved if their currently divisible tasks are divided into even smaller subtasks or if some of their currently indivisible tasks are divided.

 \begin{table}[!htbp]
 \centering\small\setlength{\tabcolsep}{4pt}
  \caption{Computational results obtained with CPLEX and BB\&R-TD algorithm over Method-R instances.} \label{tab:resultsRandom}
\resizebox{\textwidth}{!}{%
  \begin{tabular}{>{\scriptsize}rlrrrrrrrrrrrrrrc}
\hline
 & \multicolumn{3}{c}{SALBP-1} & & \multicolumn{5}{c}{BB\&R-TD} & & \multicolumn{5}{c}{CPLEX} &\\
\cline{2-4}\cline{6-10}\cline{12-16}
&	Graph	&	$c$	&	$m^*$	&	&	$m$	&	$F$	&	LE	&	LT	&	CPU	&	&	$m$	&	$F$	&	LE	&	LT	&	CPU	&	Obs.	\\
\hline
1	&	BOWMAN8	&	20	&	5	&	&	4	&	2	&	96,25	&	78	&	0,54	&	&	4	&	2	&	96,25	&	78	&	0,01	&	2	\\
2	&	GUNTHER	&	49	&	11	&	&	10	&	2	&	98,97	&	489	&	2,71	&	&	10	&	2	&	98,98	&	489	&	0,65	&	1	\\
3	&	GUNTHER	&	41	&	14	&	&	13	&	8	&	92,12	&	498	&	0,56	&	&	13	&	8	&	92,12	&	498	&	0,86	&	1	\\
4	&	JAESCHKE	&	7	&	7	&	&	6	&	3	&	95,24	&	42	&	0,56	&	&	6	&	3	&	95,24	&	42	&	0,02	&	1	\\
5	&	LUTZ2	&	11	&	49	&	&	48	&	7	&	93,18	&	525	&	397,77	&	&	49	&	0	&	90,04	&	518	&	3600,01	&	3	\\
6	&	LUTZ3	&	150	&	12	&	&	11	&	2	&	99,76	&	1650	&	1,28	&	&	11	&	2	&	99,76	&	1650	&	1,96	&	1	\\
7	&	MERTENS	&	8	&	5	&	&	4	&	3	&	100	&	32	&	0,56	&	&	4	&	3	&	100	&	32	&	0,02	&	1	\\
8	&	SAWYER30	&	30	&	12	&	&	11	&	5	&	99,7	&	329	&	0,59	&	&	11	&	5	&	99,7	&	329	&	0,52	&	1	\\
9	&	TONGE70	&	293	&	13	&	&	12	&	4	&	99,94	&	3516	&	6,41	&	&	13	&	12	&	92,47	&	3800	&	3600,01	&	3	\\
10	&	TONGE70	&	220	&	17	&	&	16	&	10	&	100	&	3520	&	0,98	&	&	16	&	5	&	99,86	&	3520	&	2,81	&	1	\\
11	&	TONGE70	&	207	&	18	&	&	17	&	5	&	99,89	&	3518	&	6,63	&	&	18	&	2	&	94,26	&	3699	&	3600,02	&	3	\\
12	&	WARNECKE	&	62	&	27	&	&	26	&	18	&	97,15	&	1610	&	1,43	&	&	26	&	26	&	97,64	&	1612	&	721,34	&	2	\\
13	&	WARNECKE	&	58	&	29	&	&	28	&	21	&	96,61	&	1621	&	4,97	&	&	28	&	14	&	96,18	&	1621	&	3600,01	&	4	\\
14	&	WARNECKE	&	54	&	31	&	&	30	&	23	&	96,98	&	1606	&	2,19	&	&	30	&	19	&	96,73	&	1620	&	184,88	&	1	\\
15	&	WEE-MAG	&	56	&	30	&	&	28	&	39	&	98,09	&	1542	&	1,33	&	&	28	&	24	&	97,13	&	1568	&	3600,01	&	3	\\
16	&	WEE-MAG	&	52	&	31	&	&	30	&	41	&	98,72	&	1552	&	2,53	&	&	30	&	21	&	97,44	&	1560	&	1235,61	&	2	\\
17	&	WEE-MAG	&	54	&	31	&	&	29	&	39	&	98,21	&	1556	&	4,62	&	&	29	&	16	&	96,74	&	1565	&	3600,01	&	3	\\
18	&	WEE-MAG	&	49	&	32	&	&	31	&	18	&	99,87	&	1519	&	1817,82	&	&	31	&	6	&	99,08	&	1519	&	2808,84	&	2	\\
19	&	WEE-MAG	&	50	&	32	&	&	31	&	38	&	99,16	&	1543	&	6,6	&	&	31	&	5	&	97,03	&	1542	&	44,2	&	1	\\
20	&	WEE-MAG	&	46	&	34	&	&	33	&	17	&	99,87	&	1516	&	284,53	&	&	34	&	12	&	96,61	&	1563	&	3600,01	&	4	\\
21	&	WEE-MAG	&	45	&	38	&	&	35	&	50	&	98,35	&	1551	&	14,81	&	&	35	&	17	&	96,25	&	1574	&	3600,02	&	4	\\
22	&	WEE-MAG	&	43	&	50	&	&	37	&	46	&	97,11	&	1570	&	1,05	&	&	36	&	39	&	99,35	&	1548	&	1815,74	&	2	\\
23	&	WEE-MAG	&	42	&	55	&	&	38	&	52	&	97,18	&	1576	&	41,49	&	&	38	&	49	&	96,99	&	1596	&	2616,98	&	1	\\
24	&	WEE-MAG	&	41	&	59	&	&	41	&	51	&	92,21	&	1662	&	1,57	&	&	41	&	48	&	92,03	&	1678	&	85,5	&	1	\\
25	&	WEE-MAG	&	35	&	60	&	&	45	&	50	&	98,35	&	1564	&	2,39	&	&	45	&	45	&	98,03	&	1575	&	395,56	&	1	\\
26	&	WEE-MAG	&	36	&	60	&	&	45	&	44	&	95,25	&	1605	&	3606,32	&	&	44	&	42	&	97,29	&	1584	&	3600,01	&	6	\\
27	&	WEE-MAG	&	37	&	60	&	&	43	&	49	&	97,3	&	1575	&	10,57	&	&	43	&	42	&	96,86	&	1591	&	123,87	&	1	\\
28	&	WEE-MAG	&	38	&	60	&	&	42	&	51	&	97,12	&	1579	&	9,75	&	&	42	&	46	&	96,8	&	1593	&	91,54	&	1	\\
29	&	WEE-MAG	&	39	&	60	&	&	42	&	50	&	94,57	&	1621	&	21,57	&	&	42	&	47	&	94,38	&	1638	&	120,93	&	1	\\
30	&	WEE-MAG	&	40	&	60	&	&	42	&	50	&	92,2	&	1662	&	3608,06	&	&	41	&	50	&	94,45	&	1640	&	3368,1	&	1	\\
31	&	WEE-MAG	&	32	&	61	&	&	52	&	45	&	92,79	&	1653	&	3614,92	&	&	52	&	37	&	92,31	&	1661	&	235,01	&	1	\\
32	&	WEE-MAG	&	33	&	61	&	&	49	&	47	&	95,61	&	1606	&	3611,47	&	&	49	&	39	&	95,11	&	1617	&	1542,53	&	5	\\
33	&	WEE-MAG	&	34	&	61	&	&	46	&	47	&	98,85	&	1557	&	1,3	&	&	46	&	39	&	98,34	&	1563	&	3600,01	&	3	\\
34	&	WEE-MAG	&	30	&	62	&	&	55	&	39	&	93,21	&	1642	&	105,66	&	&	54	&	34	&	94,63	&	1618	&	1464,25	&	2	\\
35	&	WEE-MAG	&	31	&	62	&	&	54	&	34	&	91,58	&	1665	&	3610,7	&	&	53	&	37	&	93,49	&	1642	&	1648,76	&	1	\\
36	&	WEE-MAG	&	28	&	63	&	&	62	&	32	&	88,19	&	1729	&	3614,77	&	&	62	&	11	&	86,98	&	1735	&	3600,01	&	6	\\
37	&	WEE-MAG	&	29	&	63	&	&	59	&	38	&	89,83	&	1703	&	1193,07	&	&	59	&	37	&	89,77	&	1709	&	509,63	&	1	\\
\hline
  \end{tabular}
}
  \end{table}

% SECTION -----------------------
\section{Conclusions and future research}
\label{sec:conclusions}
This paper presents a new type of assembly line balancing problem with task division, where particular tasks can potentially be divided in a limited way, even if the division induces additional time penalties. It  introduced a mathematical model of the problem in order to solve numerical instances with the commercial optimization software CPLEX and a proposed implicit enumeration algorithm based on branch, bound and remember. Preliminary results of an extensive computational study demonstrate that the approach seems promising. Possible directions for future research in order to seek a better trade-off between solution quality and computational time include the adaptation of ASALBP solution methods \citep{scholl-boysen-fliedner-2009} and metaheuristics, such as genetic algorithms and tabu search.

% ACKNOWLEDGEMENTS - - - - - - - -
% \paragraph{Acknowledgements:}
% While this research was being conducted, Carlos Alexandre X. Silva was supported by a CAPES (Brazil) grant and Les R. Foulds was partially supported by a PVNS-CAPES (Brazil) grant.

% SECTION -----------------------
% \section*{References}
\bibliographystyle{apalike}
\bibliography{cie-tdalbp}

\end{document}